\documentclass[onecolumn]{IEEEtran}
\usepackage{pgfplots}
\pgfplotsset{compat=1.5}

\usepackage{cite}
\usepackage[cmex10]{amsmath}
\usepackage{fixltx2e}
\usepackage{hyperref}
\usepackage{graphicx}
\usepackage{algorithmic}
\usepackage{algorithm}

\newtheorem{theorem}{Theorem}
\newtheorem{definition}{Definition}
\newtheorem{corollary}{Corollary}

\newcommand{\opt}{\mathrm{opt}}
\newcommand{\non}{\mathrm{non}}
 
\begin{document}

\title{Average  Convergence  Rate of Evolutionary Algorithms}
\author{Jun He and   Guangming Lin 
\thanks{This work was supported by EPSRC under Grant No. EP/I009809/1 (He), National Science Foundation of Guangdong Province under Grant No. S2013010014108 and Shenzhen Scientific Research Project under Grant No. JCYJ 20130401095559825 (Lin). }
\thanks{Jun He is with Department of Computer Science, Aberystwyth University, Aberystwyth SY23 3DB, U.K. e-mail:  jqh@aber.ac.uk.}
\thanks{Guangming Li is with Shenzhen Institute of Information Technology, China. e-mail:  lingm@sziit.com.cn‎.}  
}
\maketitle

\begin{abstract} 
In evolutionary optimization, it is important to understand how fast evolutionary algorithms converge to the optimum per generation, or their convergence rates. This paper proposes a new measure of  the  convergence rate, called the average convergence rate. It is  a normalized geometric  mean of the reduction  ratio of the fitness difference    per generation. The calculation of the average convergence rate is very simple and it is applicable for most   evolutionary algorithms   on both continuous and discrete optimization.  A theoretical study of the average convergence rate is  conducted for discrete optimization. Lower   bounds on the average convergence rate are derived. The limit of the average convergence rate  is analyzed and then the asymptotic average convergence rate is proposed. 
\end{abstract}

\section{Introduction}
Evolutionary algorithms  (EAs) belong to  iterative methods.  As iterative methods,    a fundamental question  is their  convergence rates:  how fast does an EA  converge to the optimum per generation?  
According to~\cite{ming2006convergence},   
existing results on the convergence rate of genetic algorithms  can be classified 
into two categories. The first category is related to the eigenvalues of the transition matrix associated with an EA.    A lower
bound of convergence rate is derived in \cite{suzuki1995markov} for   simple genetic algorithms by analyzing eigenvalues   of the  transition matrix. Then the work is extended in
\cite{schmitt2001importance} and it is found that the
convergence rate   is determined by the second largest
eigenvalue of the transition matrix.   The other category is
based on  Doeblin's condition.  The upper bound  on the convergence rate   is derived using   Deoblin's condition in \cite{he1999convergence}.    
As to continuous optimization,  the local convergence rate of   EAs  on  the sphere function,  quadratic convex functions and convex objective functions  are discussed in \cite{rudolph1997local,rudolph1997convergence,rudolph2013convergence}.  The research of  the convergence rate  covers various types of EAs such as isotropic algorithms \cite{teytaud2006ultimate}, gene expression programming \cite{du2010convergence},  multiobjective optimization EAs~\cite{beume2011convergence}.  The   relationship between the convergence rate and population size is investigated in \cite{jebalia2010log,teytaud2014convergence}. 

The convergence rate in previous studies~\cite{ming2006convergence,suzuki1995markov,he1999convergence,schmitt2001importance}  is based on  Markov chain theory. Suppose that an EA is modeled by a finite Markov chain with a transition matrix  $\mathbf{P}$, in which a state is a population~\cite{he2003towards}. Let $\mathbf{p}_t$ be  the probability distribution of the $t$th generation population on a population space, $\boldsymbol{\pi}$   an invariant probability distribution of $\mathbf{P}$.
Then $\mathbf{p}_t$   is called \emph{convergent} to $\boldsymbol{\pi}$ if $\lim_{t \to \infty} \parallel \mathbf{p}_t-   \boldsymbol{\pi}  \parallel=0$  where $\parallel \cdot \parallel$ is a norm; and the \emph{convergence rate} refers to the order of how fast  $\mathbf{p}_t$ converges to $\boldsymbol{\pi}$~\cite{he1999convergence}. The goal is to obtain a bound $\epsilon(t)$ such that $\parallel \mathbf{p}_t-  \boldsymbol{\pi} \parallel \le \epsilon(t)$. But to obtain a closed form of $\epsilon(t)$ often is difficult  in both theory and  practice. 

The current paper aims to seek a  convergence rate satisfying two requirements: it is easy to calculate the convergence rate  in practice while  it is possible to make a rigorous analysis in theory. Inspired from conventional iterative methods~\cite{varga2009matrix},    a new measure of the convergence rate, called the average convergence rate, is presented.  
  The paper is organized as follows: Section~\ref{secDefinition}  defines the average convergence rate.  Section~\ref{secAnalysis} establishes  lower  bounds on the average convergence rate.  Section~\ref{secConnections} discusses the connections between the average convergence rate and other performance measures. Section~\ref{secDiscussion} introduces an alternative definition of the average convergence rate if the optimal fitness value is unknown. Section~\ref{secConclusions} concludes the paper.

\section{Definition and Calculation}
\label{secDefinition}
Consider the problem of  minimizing (or maximizing) a function $  f(x)
$.  An  EA  for solving the problem is regarded as an iterative procedure (Algorithm~\ref{alg1}):  initially construct a population of solutions $\Phi_0$;    then generate a sequence of populations  $\Phi_1$, $\Phi_2$,  $\Phi_3$ and so on. This procedure is repeated until a stopping criterion is satisfied.    An archive is   used  for recording the best found solution.   
\begin{algorithm}
\caption{An EA with an archive} \label{alg1}
\begin{algorithmic}[1]
\STATE initialize a population of solutions $ \Phi_{0}$ and set $t \leftarrow 0$;
\STATE an archive records   the best solution in $\Phi_0$; 
\WHILE{the archive doesn't include an optimal solution}
\STATE generate a new population of solutions $\Phi_{t+1}$;
\STATE  update the archive if a better solution is generated; 
\STATE   $t\leftarrow t+1$;
\ENDWHILE
\end{algorithmic}
\end{algorithm}

The fitness of  population $\Phi_t$ is defined by  the  best fitness value among its individuals, denoted by $f(\Phi_t)$. Since $f(\Phi_t)$ is a random variable, we consider its mean value  $f_t:= \mathrm{E}[f(\Phi_t)]$. 
Let $f_{\opt}$ denote the optimal fitness. The \emph{fitness difference} between     $f_\opt$ and $f_t$ is $| f_{\opt}- f_t|$.
 The \textit{convergence rate} for one generation is   
\begin{align}
\left| \frac{  f_{\opt}- f_t }{  f_{\opt}- f_{t-1} }\right|.
\end{align}
Since  $|f_{\opt}- f_t| \approx |f_{\opt}- f_{t-1}|$,  calculating the above ratio  is unstable in practice. Thus a new average convergence rate for EAs is proposed in the current paper.  

\begin{definition}
Given an initial population $\Phi_0$, the \emph{ average (geometric) convergence rate  of an EA for $t$ generations} is 
\begin{align}
\label{equAverageRate}
   R(t \mid \Phi_0)&:=1- \left( \left| \frac{ f_{\opt}- f_1}{ f_{\opt}- f_{0}}\right|   \cdots  \left| \frac{  f_{\opt}- f_t }{  f_{\opt}- f_{t-1} }\right|\right)^{1/t}  \equiv 1- \left( \left| \frac{ f_{\opt}- f_t}{ f_{\opt}- f_{0}} \right|\right)^{1/t}.
\end{align}
If $f_0=f_{\opt}$, let $R(t \mid \Phi_0)=1$.  For the sake of simplicity, $R(t)$ is short for  $R(t \mid \Phi_0)$. 
\end{definition}

The rate represents a normalized geometric mean of the reduction ratio of the fitness difference  per generation. 
The larger the convergence rate, the faster the convergence.  The rate   takes the maximal value of 1 at $f_t=f_{\opt}$.   

 Inspired from  conventional iterative methods  \cite[Definition 3.1]{varga2009matrix},  the \emph{average (logarithmic) convergence rate} is defined as follows:
\begin{align}
\label{equLogRate}
R^\dagger(t):=-\frac{1}{t}\log \left| \frac{  f_{\opt}- f_t }{  f_{\opt}- f_{0} }\right|. 
\end{align}  
Formula (\ref{equLogRate}) is not adopted since its value is $+\infty$ at $f_t=f_{\opt}$. But in most cases,   average geometric and logarithmic convergence rates are almost the same.  Let $\alpha_t:=|f_{\opt}- f_t|/ |f_{\opt}- f_{t-1}|$. Usually $\alpha_t \approx 1$ and  $(\alpha_1 \cdots \alpha_t )^{1/t} \approx 1 
$, then $-\log (\alpha_1 \cdots \alpha_t )^{1/t} \div (1-(\alpha_1 \cdots \alpha_t )^{1/t}) \approx 1$.

In practice, the average convergence rate is calculated as follows: given   $f(x)$ with $f_{\opt}$ known in advance,  
\begin{enumerate}
 \item Run an EA for $T$ times ($T \gg 1$).
 
 \item Then   calculate the mean fitness value $f_t$  as follows,
\begin{equation}
\label{equft}
\frac{1}{T} \left(f(\Phi^{[1]}_t) +\cdots +f(\Phi^{[T]}_t) \right),
\end{equation}
where $f(\Phi^{[k]}_t)$ denotes the fitness $f(\Phi_t)$ at the $k$th run.  The law of large numbers guarantees  (\ref{equft}) approximating to the mean fitness value $f_t =\mathrm{E}(f(\Phi_t))$ when   $T$ tends towards $+\infty$.

\item Finally,  calculate $R(t)$ according to formula (\ref{equAverageRate}).
\end{enumerate}

\begin{figure}[ht]
\begin{center}
\begin{tikzpicture} 
\begin{axis}[width=8cm,height=5cm,
scaled ticks=false,
legend style={draw=none},
xmin=0,
ymin=0,
ymax=0.011,
xtick={0, 500,1000,1500},
ytick={0, 0.0025,0.005,0.0075,0.01},
xlabel={$t$},  
]
\addplot[red] table {f10-geom-rate-MEA.dat}; 
\addplot[green,dashed] table {f10-geom-rate-FEP.dat}; 
    \legend{MEA,FEP}
\end{axis} 
\end{tikzpicture}
\end{center}
\caption{A comparison of the average convergence rates   $R(t)$ of MEA and FEP on Ackley's function.}
\label{figAckley}
\end{figure}
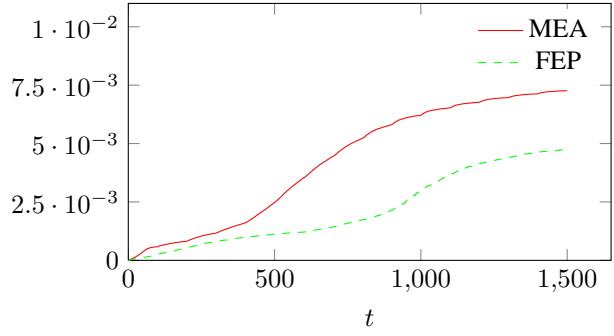

The calculation is applicable for most EAs   on both continuous and discrete optimization. 
We take an example to illustrate   the average convergence rate. Consider the problem of minimizing   Ackley's function:
\begin{align}
f(x) =&  -20 \exp \{-0.2 [\sum^n_{i=1} (x_i+e)^2 /n]^{\frac{1}{2}}\}  -\exp[\sum^n_{i=1} \cos(2 \pi x_i+2 \pi e) /n ]  +20 +e,
\end{align}
where   $x_i  \in [-32-e, 32-e], i=1, \cdots, n$. The optimum is $(-e, -e,\cdots)$ and  $f_{\opt}=0$.  
We compare the Multi-grid EA (MEA)~\cite{he2009mixed}  with the Fast Evolutionary Programming (FEP)~\cite{yao1999evolutionary}  under the same experiment setting  (where $n$ is 30 and population size is 100).     Run the two EAs for 1500 generations and 100 times. Calculate $f_t$ according to (\ref{equft}) and then $R(t)$ according to (\ref{equAverageRate}).
Fig.~\ref{figAckley} illustrates the convergence rates of MEA and FEP.  
 
The average convergence rate is different from the progress rate such as  $|f_t - f_{\opt}|$ or logarithmic rate $\log |f_t-f_{\opt}|$ used in \cite{salomon1998evolutionary}. The progress rate measures the fitness change; but the convergence rate measures the   rate of the fitness change. We demonstrate this difference by an example. Let $g(x)=100 f(x)$. In terms of $|f_t - f_{\opt}|$, the progress rate  on $g(x)$ is 100 times that on $f(x)$. In terms of $\log |f_t - f_{\opt}|$, the progress rate  on $g(x)$ is $1+2/ \log |f_t - f_{\opt}|$ times that on $f(x)$. However, the average convergence rate is the same on both $f(x)$ and $g(x)$.

\section{Analysis for Discrete  Optimization}
\label{secAnalysis}
Looking at Fig.~\ref{figAckley} again,  two  questions may be raised: what is the lower bound or upper bound on $R(t)$? Does $R(t)$ converge   or not?
For discrete optimization,  a  theoretical answer is provided to these questions in this section.  For continuous optimization, its analysis  is left for future research.    

In the rest of the paper, we analyze EAs for discrete optimization and assume that their genetic operators do not change with time. Such an EA  can be modeled by a homogeneous Markov chain~\cite{he2003towards} with  transition probabilities 
$
\Pr(X,Y):=\Pr(\Phi_{t+1}= Y \mid  \Phi_t = X),   X , Y \in S,
$
where  populations $X, Y$  denote   states of $\Phi_t$ and $S$ denotes the  set of populations (called the \emph{population space}).  
Let $\mathbf{P}$ denote  the transition matrix with entries $\Pr(X,Y)$.

A population is called \emph{optimal} if it includes an optimal solution; otherwise called \emph{non-optimal}. Let $S_{\opt}$ denote   the set of  optimal populations, and $S_{\non}= S \setminus S_{\opt}$. Because of   the stopping criterion,  the  optimal set is always absorbing,  
\begin{align}
\Pr (\Phi_{t+1} \in S_{\non}  \mid \Phi_t  \in S_{\opt})=0.
\end{align}

Transition matrix $\mathbf{P} $  can be split into four parts: 
\begin{equation}
\label{equCanonicalForm} 
\mathbf{P} = \bordermatrix{
%\begin{pmatrix}{
~& S_{\opt} & S_{\non} \cr
S_{\opt}&\mathbf{A} & \mathbf{{  O}} \cr
S_{\non}&  \mathbf{B}  & \mathbf{Q} \cr
%\end{pmatrix},
}
\end{equation}
where  
$\mathbf{A}$ is a submatrix representing    probability transitions among  optimal  states;   $\mathbf{O}$ a submatrix for probability transitions  from optimal  states  to non-optimal ones,  of which all entries take the value of zero; $ \mathbf{B}$ a submatrix denoting    probability transitions from non-optimal  states to optimal ones; and $ \mathbf{Q}$ a    submatrix for    probability transitions among non-optimal  states.  
 
Since  $\Phi_t$ is a random variable, we investigate the probability  distribution of $\Phi_t$ instead of $\Phi_t$ itself. Let $q_t(X)$ denote the probability of $\Phi_t$ at a non-optimal state $X$, 
$
q_t(X): =\Pr(\Phi_t=X).
$ Let  vector $(X_1, X_2, \cdots )$ represent all non-optimal states and vector
 $\mathbf{q}^T_t$ denote   the probability distribution of  $\Phi_t$ in the non-optimal set, where
$
\mathbf{q}_t:= (q_t(X_1), q_t(X_2), \cdots   )^T.
$ Here notation  $\mathbf{q}$ is a column vector and $\mathbf{q}^T$ the row column with the transpose operation. For the  initial probability distribution, $\mathbf{q}_0 \ge \mathbf{0}$   where $\mathbf{0}=(0, 0, \cdots)^T$. Only  when the initial population is chosen from the optimal set, $\mathbf{q}_0=0$.

Consider   probability transitions among non-optimal states only, which can be represented by   matrix iteration
\begin{equation}
\label{equMatrixIteration}
\mathbf{q}^T_{t}= \mathbf{q}^T_{t-1}   \mathbf{Q} =\mathbf{q}^T_0  \mathbf{Q}^t.  
\end{equation}

\begin{definition} 
An EA  is called \emph{convergent} if
$ 
\lim_{t \to +\infty}  \mathbf{q}_t = \mathbf{0}$ for any $\mathbf{q}_0$ or $ 
\lim_{t \to +\infty}  \mathbf{Q}^t = \mathbf{O}$. It is equivalent to saying that the probability of finding an optimal solution  is $1$ as $t$ tends towards $+\infty$. 
\end{definition}

The mean fitness value $f_t$  is given as follows: 
\begin{align}
\label{equMeanFitness}
f_t :  = \mathrm{E}[f(\Phi_t)]=\sum_{X \in S } f(X) \Pr(\Phi_t =X). 
\end{align}
Then  it follows  
\begin{align}
\label{equFitnessValue}
   f_{\opt}-f_t   &= \sum_{X \in S_{\non}}   (f(X)-f_{\opt}) q_t(X) .
\end{align}

Let    vector  $\mathbf{f}:=(f(X_1), f(X_2), \cdots)^T$  denote the fitness values of all non-optimal populations $(X_1, X_2, \cdots)$. Then (\ref{equFitnessValue}) can be rewritten in a vector form
\begin{align}
\label{equDifference}
  f_{\opt}-f_t  
=   \mathbf{q}^T_t \cdot (f_{\opt} \mathbf{1}-\mathbf{f}),
\end{align}
where $\cdot$ denotes the vector product   and  $\mathbf{1}=(1,1, \cdots)^T$.  

For a  vector $\mathbf{v}$, denote 
\begin{align}
\parallel \mathbf{v}^T  \parallel:= |\mathbf{v}^T \cdot (f_{\opt} \mathbf{1}-\mathbf{f})|.
\end{align}
Since $\parallel \mathbf{v} \parallel =0$ iff $\mathbf{v}=\mathbf{0}$; $\parallel a \mathbf{v}\parallel =| a |\parallel \mathbf{v} \parallel$ and $\parallel \mathbf{v}_1 +\mathbf{v}_2 \parallel \ge \parallel \mathbf{v}_1 \parallel +\parallel \mathbf{v}_2 \parallel$, thus $\parallel \mathbf{v} \parallel$ is a vector norm. 
 For a matrix $\mathbf{M}$, let $\parallel \mathbf{M} \parallel$ be the induced matrix norm, given by
 \begin{align}
 \parallel \mathbf{M} \parallel =\sup \left\{ \frac{\parallel \mathbf{v}^T \mathbf{M} \parallel}{\parallel \mathbf{v}^T \parallel}: \mathbf{v} \neq \mathbf{0} \right\}.
 \end{align}

Using the above Markov chain model,  we  are able to estimate lower   bounds on  the  average  convergence rate.

\begin{theorem} 
\label{theAverageRate}
Let $\mathbf{Q}$ be the transition submatrix associated with a convergent EA. For any $\mathbf{q}_0 \neq \mathbf{0}$,
\begin{enumerate}
\item The   average  convergence rate for $t$ iterations  is lower-bounded by
\begin{align}
\label{equFirstConclusion}
   R(t)
      \ge& 1- \parallel \mathbf{Q}^t \parallel^{1/t}  .
\end{align}

\item  The   limit of the average  convergence rate  for $t$ generations  is lower-bounded by
 \begin{align}
 \label{equSecondConclusion}
   \lim_{t \to +\infty}  R(t)
   \ge   1-\rho(\mathbf{Q}),
\end{align} 
where $\rho(\mathbf{Q})$  is the spectral radius   (i.e., the supremum among the absolute values of all eigenvalues of $\mathbf{Q}$). 

\item Under random initialization (that is, $\Pr(\Phi_0=X) >0$  for any $X \in S_{\non}$ or $\mathbf{q}_0 > \mathbf{0}$),    it holds
\begin{align}
\label{equThirdConclusion}
   \lim_{t \to +\infty}  R(t)
 =   1-\rho(\mathbf{Q}).
\end{align}

\item Under particular initialization (that is, set\footnote{For vector $\mathbf{v}=(v_1, v_2, \cdots)$, denote $|\mathbf{v}|:=\sum_i |v_i|$.} $\mathbf{q}_0= \mathbf{v}/|\mathbf{v}|$ where $\mathbf{v}$ is an eigenvector corresponding to the eigenvalue  $\rho(\mathbf{Q})$  with $\mathbf{v}\ge \mathbf{0}$ but $\mathbf{v} \neq  \mathbf{0}$. The existence of such a $\mathbf{v}$  is given in the proof),  it holds  for all $t\ge 1$,
 \begin{align}
 \label{equFourthConclusion}
  R(t)
 =   1-\rho(\mathbf{Q}).
\end{align}
\end{enumerate} 
\end{theorem}

\begin{IEEEproof}
1)  From (\ref{equMatrixIteration}):
$  \mathbf{q}^T_t  = \mathbf{q}^T_0   \mathbf{Q}^t,$
 we have 
 \begin{align}
& \frac{| f_{\opt}- f_t|}{ |f_{\opt}- f_0|} 
  =\frac{\parallel \mathbf{q}^T_t \parallel}{\parallel \mathbf{q}^T_0 \parallel} =\frac{\parallel  \mathbf{q}^T_0  \mathbf{Q}^t \parallel}{\parallel \mathbf{q}^T_0 \parallel} \le \frac{\parallel  \mathbf{q}^T_0 \parallel \parallel \mathbf{Q}^t \parallel}{\parallel \mathbf{q}^T_0 \parallel}  
   = \parallel \mathbf{Q}^t \parallel   .
\end{align}

Hence \begin{align}
\label{equFirstResult}
   1- \left| \frac{ f_{\opt}- f_t}{ f_{\opt}- f_0}\right|^{1/t}  
  \ge    1-\parallel \mathbf{Q}^t \parallel^{1/t} ,
\end{align}
which proves the first conclusion.

2) According to   Gelfand's spectral radius formula \cite[p.619]{meyer2000matrix},  we get
\begin{align}
\label{equSecondResult}
  \lim_{t \to +\infty} \parallel \mathbf{Q}^t \parallel^{1/t}    =  \rho(\mathbf{Q}).
\end{align}
The second conclusion follows by combining (\ref{equSecondResult}) with (\ref{equFirstConclusion}).

 3) Since  $\mathbf{Q}\ge 0$, according to  the extension of Perron-Frobenius' theorems to non-negative matrices~\cite[pp. 670]{meyer2000matrix}, $\rho(\mathbf{Q})$ is an eigenvalue of $\mathbf{Q}$. There exists an eigenvector   $\mathbf{v}$  corresponding to $\rho(\mathbf{Q})$  such that  $\mathbf{v}\ge \mathbf{0}$ but $\mathbf{v} \neq \mathbf{0}$. In particular,
\begin{align}
\label{equEigenvector}
 \rho(\mathbf{Q})\mathbf{v}^T  = \mathbf{v}^T \mathbf{Q}.
\end{align}

Let $\max(\mathbf{v})$ denote the maximum value of the  entries in  vector $\mathbf{v}$.  Due to random initialization,   $\mathbf{q}_0 >0$. Let $\min(\mathbf{q}_0)$ denote the minimum  value of the  entries in vector $\mathbf{q}_0$.      Set  
\begin{align}
\label{equU}
 \mathbf{u} = \frac{\min (\mathbf{q}_0)}{\max( \mathbf{v})} \mathbf{v}.
\end{align}
From (\ref{equEigenvector}), we get
\begin{align}
\rho(\mathbf{Q})  \mathbf{u}^T  =  \mathbf{u}^T   \mathbf{Q}.
\end{align}
Thus vector $ \mathbf{u} $ is an eigenvector of $\rho(\mathbf{Q})$.

Let
$ %\begin{align}
\mathbf{w} =\mathbf{q}_0- \mathbf{u}.
$  
Then  from (\ref{equU}), we know
$\mathbf{w}\ge 0$.
 Since
$ %\begin{align}
\mathbf{q}_0 = \mathbf{u} +\mathbf{w},
$ %\end{align}
  $\mathbf{w}\ge 0$ and $\mathbf{Q} \ge 0$, we deduce that
\begin{align}
 & \mathbf{q}^T_t   = \mathbf{q}^T_0 \mathbf{Q}^t  = ( \mathbf{u} +\mathbf{w})^T\mathbf{Q}^t  \ge  \mathbf{u}^T \mathbf{Q}^t    = \rho(\mathbf{Q})^t\mathbf{u}^T  .
\end{align}
It follows that
 \begin{align}
&\left|   \frac{ f_{\opt}- f_t}{ f_{\opt}- f_0} \right| = \left| \frac{\mathbf{q}^T_t \cdot (f_{\opt} \mathbf{1}-\mathbf{f})}{ f_{\opt}- f_0}\right| \ge   \left|    \frac{ \rho(\mathbf{Q})^t    \mathbf{u} ^T    \cdot (f_{\opt} \mathbf{1}-\mathbf{f} )}{ \mathbf{q}^T_0 \cdot (f_{\opt} \mathbf{1}-\mathbf{f})} \right|.
\end{align}
 \begin{align}
\left| \frac{ f_{\opt}- f_t}{ f_{\opt}- f_0} \right|^{1/t}  
\ge \rho(\mathbf{Q}) \left|\frac{  \mathbf{u} ^T    \cdot (f_{\opt} \mathbf{1}-\mathbf{f} )}{ \mathbf{q}^T_0 \cdot (f_{\opt} \mathbf{1}-\mathbf{f})}  \right|^{1/t}.
\end{align}

Since both $|\mathbf{u} ^T    \cdot (f_{\opt} \mathbf{1}-\mathbf{f} ) |$ and  $|\mathbf{q}^T_0 \cdot (f_{\opt} \mathbf{1}-\mathbf{f})|  $ are independent of $t$, we let $t \to +\infty$ and get \begin{align}
\lim_{t \rightarrow + \infty} \left|\frac{  \mathbf{u} ^T    \cdot (f_{\opt} \mathbf{1}-\mathbf{f} )}{ \mathbf{q}^T_0 \cdot (f_{\opt} \mathbf{1}-\mathbf{f})}  \right|^{1/t} = 1,
\end{align}
 then we get 
\begin{align}
\lim_{t \rightarrow + \infty} \left| \frac{ f_{\opt}- f_t}{ f_{\opt}- f_0} \right|^{1/t}  
\ge    \rho(\mathbf{Q}).
\end{align}
\begin{align}
\label{equThirdInequality}
\lim_{t \rightarrow + \infty} R(t) =1-\lim_{t \rightarrow + \infty} \left| \frac{ f_{\opt}- f_t}{ f_{\opt}- f_0} \right|^{1/t}  
\le  1-  \rho(\mathbf{Q}).
\end{align}
The third conclusion follows by combining (\ref{equThirdInequality}) with (\ref{equSecondConclusion}).

 4) Set  $\mathbf{q}_0= \mathbf{v}/\sum_i v_i$ where $\mathbf{v}$ is given in Step 3. Then $\mathbf{q}_0$ is an eigenvector corresponding to the eigenvalue $\rho(\mathbf{Q})$ such that  $\rho(\mathbf{Q}) \mathbf{q}^T_0 =    \mathbf{q}^T_0 \mathbf{Q}$. 
 From (\ref{equMatrixIteration}): $\mathbf{q}^T_{t}= \mathbf{q}^T_{t-1}   \mathbf{Q}$,  we get  
 \begin{align*}
   &\frac{ f_{\opt}- f_t}{ f_{\opt}- f_0}  = \frac{\mathbf{q}^T_t \cdot (f_{\opt} \mathbf{1}-\mathbf{f})}{ f_{\opt}- f_0} =       \frac{ \rho(\mathbf{Q})^t    \mathbf{q}^T_0    \cdot (f_{\opt} \mathbf{1}-\mathbf{f} )}{ \mathbf{q}^T_0 \cdot (f_{\opt} \mathbf{1}-\mathbf{f})} .
\end{align*}
Thus we have for any $t \ge 1$ 
\begin{align}
\left| \frac{ f_{\opt}- f_t}{ f_{\opt}- f_0} \right|^{1/t}  
=  \rho(\mathbf{Q}),
\end{align}
then $R(t)=1-\rho(\mathbf{Q})$ which gives the fourth conclusion.
\end{IEEEproof}
 
The above theorem provides lower bounds on the average convergence rate. Furthermore, it reveals that $R(t)$ converges to $1-\rho(\mathbf{Q})$    under  random initialization and $R(t)=1-\rho(\mathbf{Q})$ for any $t \ge 1$ under particular initialization.  
Similar to conventional iterative methods \cite[pp. 73]{varga2009matrix},  we call 
 $1-\rho(\mathbf{Q})$   the \emph{asymptotic average convergence rate} of an EA, denoted by $R_{\infty}$.   According to (\ref{equThirdConclusion}),  its value can be approximately calculated as follows: under random initialization,   $R(t)$ approximates to $1-\rho(\mathbf{Q})$ if $t$ is sufficiently large.  Note that this definition is different from another asymptotic convergence rate, given by $-\log \rho(\mathbf{Q})$ in~\cite{he2012pure}.  In most cases, the two rates are almost the same since usually $\rho(\mathbf{Q})\approx 1$ and then $-\log \rho(\mathbf{Q})\div(1-\rho(\mathbf{Q})) \approx 1$. Since $1-\rho(\mathbf{Q})$ is independent of  $t$ and initialization, hence  using asymptotic average convergence rate is  convenient for comparing two EAs, for example, to analyze  mixed strategy EAs~\cite{he2012pure}.

\section{Connections}
\label{secConnections}

The average convergence rate is   different from other performance measures of EAs:   the expected hitting time is the total number of generations for obtaining an optimal solution~\cite{he2003towards}; and  fixed budget analysis  focuses on the performance of EAs within fixed budget computation~\cite{jansen2014performance}. However, there are some interesting connections between them.

There exists a link between the asymptotic average convergence rate and the hitting time. Let $m(X)$ be the expected number of generations for a convergent EA to hit $S_{\opt}$ when starting from   state $X$ (called the \emph{expected  hitting time}).  Denote $\mathbf{m} :=( m(X_1), m(X_2), \cdots)^T$ where   $(X_1, X_2, \cdots )$ represent all non-optimal states. 
\begin{theorem}
\label{theLink}
Let $\mathbf{Q}$ be the transition submatrix associated with a convergent EA. 
 Then 
$1/R_{\infty}$ is not more than $\parallel \mathbf{m} \parallel_{\infty}:=\max \{ m(X); X \in S_{\non}\}.$
\end{theorem} 

\begin{IEEEproof} 
  According to the fundamental matrix theorem~\cite[Theorem 11.5]{grinstead1997introduction}, $\mathbf{m}=(\mathbf{I}-\mathbf{Q})^{-1} \mathbf{1}$, where $\mathbf{I}$ is  the unit matrix. Then 
\begin{align}
\parallel \mathbf{m} \parallel_{\infty} &= \parallel (\mathbf{I}-\mathbf{Q})^{-1} \mathbf{1}\parallel_{\infty}=\parallel (\mathbf{I}-\mathbf{Q})^{-1}  \parallel_{\infty} \ge \rho((\mathbf{I}-\mathbf{Q})^{-1}) = (1-\rho(\mathbf{Q}))^{-1},
\end{align}
where the last equality takes use of a fact: $(1-\rho(\mathbf{Q}))^{-1}$ is an eigenvalue and spectral radius of $(\mathbf{I}-\mathbf{Q})^{-1} $. 
\end{IEEEproof}

The above theorem shows that $1/R_{\infty}$ is a lower bound on the expected hitting time.

Following Theorem~\ref{theAverageRate}, a straightforward connection can be established between the spectral radius $\rho(\mathbf{Q})$ and the progress rate $|f_{\opt}-f_t|$.

\begin{corollary} 
\label{corLink}Let $\mathbf{Q}$ be the transition submatrix associated with a convergent EA. 
\begin{enumerate} 
 \item Under random initialization (that is  $\mathbf{q}_0 > \mathbf{0}$),    it holds
 \begin{align}
 \label{equLink2}
  \lim_{t \to +\infty} \frac{ |f_{\opt}-f_t| }{
   \rho(\mathbf{Q})^t  |f_{\opt}-f_0|}=1.
\end{align}

\item Under particular initialization (that is, set $\mathbf{q}_0= \mathbf{v}/|\mathbf{v}|$ where $\mathbf{v}$ is an eigenvector corresponding to the eigenvalue  $\rho(\mathbf{Q})$  with $\mathbf{v}\ge \mathbf{0}$ but $\mathbf{v} \neq  \mathbf{0}$),  it holds  for all $t\ge 1$,
 \begin{align}
 \label{equLink3}
 \frac{ |f_{\opt}-f_t| }{
    \rho(\mathbf{Q})^t |f_{\opt}-f_0|}=1.
\end{align}
\end{enumerate} 
\end{corollary}

The exponential decay, $\rho(\mathbf{Q})^t |f_{\opt}-f_0|$, provides a theoretical prediction for the trend of $|f_{\opt}-f_t| $. The   corollary confirms that the spectral radius $\rho(\mathbf{Q})$ plays an important role on the convergence rate~\cite{schmitt2001importance}.

We explain the theoretical results by a simple example. Consider  a (1+1) EA for maximizing the OneMax function $|\mathbf{x}|$ where $\mathbf{x}=(s_1, \cdots, s_n) \in\{ 0,1, \}^n$. 

\begin{algorithm}
\caption{A (1+1) elitits EA} \label{alg2}
\begin{algorithmic} 
\STATE    \emph{Onebit Mutation}:  choose a bit of   $\Phi_t$ (one individual) uniformly at random and flip it. Let $\Psi_t$ denote the  child.

\STATE  \emph{Elitist Selection:} if $f(\Psi_t) >f(\Phi_t)$, then let $\Phi_{t+1} \leftarrow \Psi_t$; otherwise $\Phi_{t+1} \leftarrow\Phi_{t}$.
\end{algorithmic}
\end{algorithm}
 
Denote subset $S_k:=\{ \mathbf{x}: |\mathbf{x}| = n-k\}$ where $k=0, \cdots, n$.  Transition probabilities satisfy that $\Pr(\Phi_{t+1} \in S_{k-1} \mid \Phi_{t} \in S_k) =\frac{k}{n}$ and $\Pr(\Phi_{t+1} \in S_k \mid \Phi_{t} \in S_k) =1-\frac{k}{n}$. Writing them in  matrix $\mathbf{P}$ (where submatrix $\mathbf{Q}$ in the bold font):
\begin{equation}
 \begin{pmatrix} 
 1 & 0 & 0& \cdots & 0&0 & 0\\
  \frac{1}{n} &  \mathbf{1-\frac{1}{n}}  & \mathbf{0}& \cdots &   \mathbf{0} &   \mathbf{0}  &   \mathbf{0} \\[0.3em]
   0&   \mathbf{\frac{2}{n} }&   \mathbf{1- \frac{2}{n}} & \cdots & \mathbf{0}&  \mathbf{0 }& \mathbf{ 0}\\
 \vdots& \vdots & \vdots & \vdots& \vdots & \vdots &\vdots  \\
0 &  \mathbf{0}&  \mathbf{0}& \cdots &  \mathbf{\frac{n-1}{n}} &  \mathbf{1-\frac{n-1}{n}} &  \mathbf{0}\\
 0  &  \mathbf{0}&  \mathbf{0}& \cdots & \mathbf{0} &   \mathbf{1} &  \mathbf{0}\\
 \end{pmatrix}.
\end{equation}
 
The spectral radius $\rho{(\mathbf{Q})}=1-\frac{1}{n}$ and  the asymptotic average convergence rate $R_{\infty}=\frac{1}{n	}$. Notice that $1/R_{\infty}(=n)$  is less than the expected   hitting time  $(=n (1+\frac{1}{2}+ \cdots +\frac{1}{n}))$. 

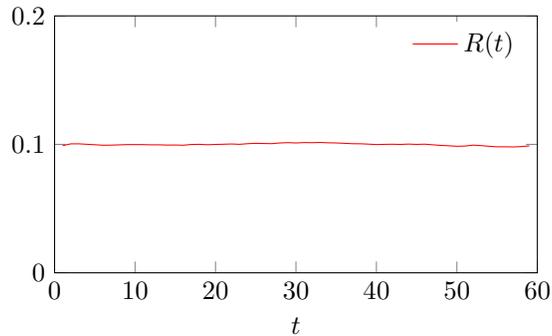
\begin{figure}[ht]
\begin{center}
\begin{tikzpicture} 
\begin{axis}[width=8cm,height=5cm,
scaled ticks=false,
legend style={draw=none},
xlabel={$t$}, 
xmin=0,
xmax=60,
ymin=0,
ymax=0.2,
ytick={0,   0.1,0.2},  
]
\addplot[red] table {geom-rate.dat}; 
    \legend{ $R(t)$}
\end{axis} 
\end{tikzpicture}
\end{center}
\caption{$R(t)$ approximates $  0.1$ for the (1+1) EA on the OneMax function with $n=10$.}
\label{figOneMax}
\end{figure}

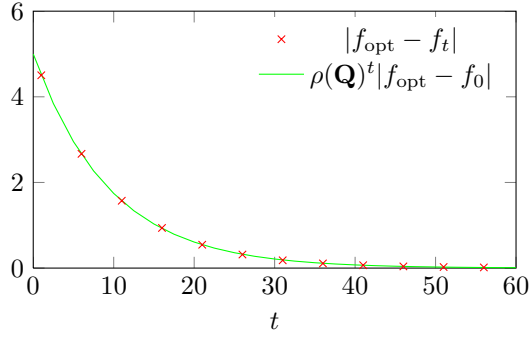
\begin{figure}[ht]
\begin{center}
\begin{tikzpicture}
\begin{axis}[ width=8cm,height=5cm,
legend style={draw=none},
xlabel={$t$}, 
xmin=0,
xmax=60,
ymin=0, 
ymax=6,
] 
\addplot[red,only marks, mark=x] table {difference.dat};
\addplot[green,domain=0:60 ]   {5*0.9^x}; 
\legend{ $| f_{\opt}-f_t|$,$\rho(\mathbf{Q})^t |f_{\opt}-f_0|$}
\end{axis} 
\end{tikzpicture}
\end{center}
\caption{A comparison of  the theoretical prediction $\rho(\mathbf{Q})^t |f_{\opt}-f_0|$ and the computational result $|f_{\opt}-f_t|$ for the (1+1) EA on the OneMax function with $n=10$, $f_0=5$, $f_{\opt}=10$ and $\rho(\mathbf{Q})=0.9$.}
\label{figOneMax2}
\end{figure}

In the OneMax function, set $n=10$, and then $\rho{(\mathbf{Q})}=0.9$ and $ R_{\infty}=0.1$. Choose $\Phi_0$ uniformly at random, run the   (1+1) EA for 50  generations and 2000 times, and then calculate  $f_t$ according to (\ref{equft}) and  $R(t)$ according to formula (\ref{equAverageRate}). Since $\Phi_0$ is chosen uniformly at random, $f_0 \approx 5$. Fig.~\ref{figOneMax} demonstrates that $R(t)$ approximates $0.1(=R_{\infty})$.
Fig.~\ref{figOneMax2} shows   that the theoretical exponential decay, $\rho(\mathbf{Q})^t |f_{\opt}-f_0|$, and the  computational progress rate,  $|f_{\opt}-f_t| $, coincide perfectly.

\section{Alternative Rate}
\label{secDiscussion}
So far the calculation of the average convergence rate needs the information about $f_{\opt}$. However this requirement is very strong. Here we introduce an alternative  average convergence rate without  knowing $f_{\opt}$, which is given as below,
\begin{align}
\label{equNewAverageRate}
R^\ddagger(t):=1-\left|\frac{f_{t+\delta t}-f_t}{f_t -f_{t-\delta t}}\right|^{1/\delta t},
\end{align}
where $\delta t$ is an appropriate time interval.  Its value relies on an EA and a problem. 

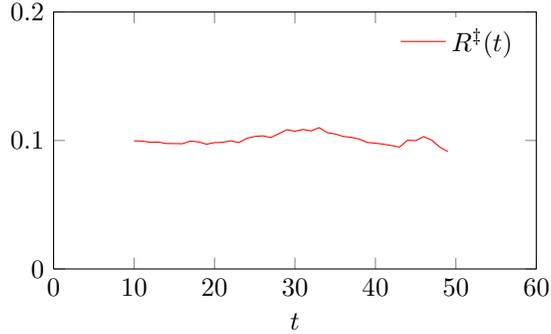
\begin{figure}[ht]
\begin{center}
\begin{tikzpicture} 
\begin{axis}[width=8cm,height=5cm,
scaled ticks=false,
legend style={draw=none},
xlabel={$t$}, 
xmin=0,
xmax=60,
ymin=0,
ymax=0.2,
ytick={0,   0.1,0.2}, 
]
\addplot[red] table {rate.dat};
    \legend{$R^{\ddagger}(t)$}
\end{axis} 
\end{tikzpicture}
\end{center}
\caption{ $R^\ddagger(t)$ approximates  $ 0.1$  for the (1+1) EA on the OneMax function with $n=10$. }
\label{figOneMax3}
\end{figure}

For the (1+1) EA on the OneMax function with $n=10$, we set $\delta t=10$. Choose $\Phi_0$ uniformly at random, run the   (1+1) EA for 60  generations and 2000 times, and then calculate  $f_t$ according to (\ref{equft}) and  $R^\ddagger(t)$ according to formula (\ref{equNewAverageRate}). Due to $\delta t=10$, $R^\ddagger(t)$ has no value for $t <10$ and $t > 50$ according to formula (\ref{equNewAverageRate}). Fig.~\ref{figOneMax3} demonstrates that $R^\ddagger(t)$ approximates $0.1 (=1-\rho(\mathbf{Q}))$. But the calculation of $R^\ddagger(t)$  is not as stable as that of $R(t)$  in practice.  

The above average convergence rate converges to $1-\rho(\mathbf{Q})$ but under stronger conditions than that in Theorem~\ref{theAverageRate}. 
\begin{theorem} 
\label{theNewAverageRate}
Let $\mathbf{Q}$ be the transition submatrix associated with a convergent EA.  
\begin{enumerate}
 
\item Under particular initialization (that is, set $\mathbf{q}_0= \mathbf{v}/|\mathbf{v}|$ where $\mathbf{v}$ is an eigenvector corresponding to the eigenvalue  $\rho(\mathbf{Q})$  with $\mathbf{v}\ge \mathbf{0}$ but $\mathbf{v} \neq  \mathbf{0}$),  it holds  for all $t\ge 1$,
 \begin{align}
 \label{equNewFirstConclusion}
  R^\ddagger(t)
 =   1-\rho(\mathbf{Q}).
\end{align}

\item Under random initialization (that is  $\mathbf{q}_0 > \mathbf{0}$), choose an appropriate $\delta t$ such that  $\mathbf{g}:= (\mathbf{I}-\mathbf{Q}^{\delta t}) (f_{\opt} \mathbf{1}-\mathbf{f}) > \mathbf{0}$ for a maximization problem  (or  $\mathbf{g}<\mathbf{0} $ for a minimization problem)\footnote{It  is always true for a large time interval $\delta t$ since   $ \lim_{\delta t\to +\infty}(\mathbf{I}- \mathbf{Q}^{\delta t})=\mathbf{I}$ and  $f_{\opt} \mathbf{1}-\mathbf{f}> \mathbf{0}$ for a maximization problem (or  $f_{\opt} \mathbf{1}-\mathbf{f} < \mathbf{0}$ for a minimization problem).  }.   If $\mathbf{Q} $ is positive\footnote{The condition 
of positive $\mathbf{Q}$ could be relaxed to non-negative $\mathbf{Q}$ if taking a similar argument to   the extension of Perron-Frobenius' theorems to non-negative matrices~\cite[pp. 670]{meyer2000matrix}.}, then it holds
\begin{align}
\label{equNewSecondConclusion}
   \lim_{t \to +\infty}  R^\ddagger(t)
 =   1-\rho(\mathbf{Q}).
\end{align}
\end{enumerate} 
\end{theorem}

\begin{IEEEproof}
 From (\ref{equMatrixIteration}): $\mathbf{q}^T_{t}= \mathbf{q}^T_{t-1}   \mathbf{Q}$ and (\ref{equDifference}),  we get  
 \begin{align}
   & f_{t+\delta t}-f_t  =  f_{t+\delta t}-f_{\opt}+f_{\opt}-f_t   =  \mathbf{q}^T_t \cdot (f_{\opt} \mathbf{1}-\mathbf{f}) - \mathbf{q}^T_{t+\delta t} \cdot (f_{\opt} \mathbf{1}-\mathbf{f}) \nonumber \\
      & =  \mathbf{q}^T_t \cdot (f_{\opt} \mathbf{1}-\mathbf{f}) - \mathbf{q}^T_{t} \mathbf{Q}^{\delta t} (f_{\opt} \mathbf{1}-\mathbf{f})  =  \mathbf{q}^T_t \cdot \mathbf{g} \label{equNewQ}.
\end{align}

1)  Since $\mathbf{q}_0$ is an eigenvector corresponding to the eigenvalue $\rho(\mathbf{Q})$ such that  $\rho(\mathbf{Q}) \mathbf{q}^T_0 =    \mathbf{q}^T_0 \mathbf{Q}$. 
 From  (\ref{equNewQ}) and (\ref{equMatrixIteration}): $\mathbf{q}^T_{t}= \mathbf{q}^T_{t-1}    \mathbf{Q}$,  we get  
 \begin{align}
   &\left|\frac{f_{t+\delta t}-f_t}{f_t -f_{t-\delta t}}\right|^{1/\delta t}   =   \left|\frac{\mathbf{q}^T_t \cdot \mathbf{g}}{\mathbf{q}^T_{t-\delta t} \cdot \mathbf{g}}\right|^{1/\delta t} =\left|\frac{\mathbf{q}^T_0 \mathbf{Q}^t \mathbf{g}}{\mathbf{q}^T_{0} \mathbf{Q}^{t-\delta t}\mathbf{g}}\right|^{1/\delta t}  =  \left| \frac{\rho(\mathbf{Q})^t }{\rho(\mathbf{Q})^{t-\delta t}} \times \frac{\mathbf{q}^T_0 \cdot \mathbf{g}}{ \mathbf{q}^T_{0} \cdot \mathbf{g}}\right|^{1/\delta t} =\rho(\mathbf{Q}).
\end{align}
Then $R^\ddagger(t)=1-\rho(\mathbf{Q})$ which gives the first conclusion.

2) Without loss of the generality, consider $\mathbf{g}>\mathbf{0}$.  
Since
\begin{align}
&\frac{f_{t+\delta t}-f_t}{f_t -f_{t-\delta t}}   =   \frac{\mathbf{q}^T_t \cdot \mathbf{g}}{\mathbf{q}^T_{t-\delta t} \cdot \mathbf{g}}= \frac{\mathbf{q}^T_{t-\delta t} \mathbf{Q}^{\delta t}   \mathbf{g}}{\mathbf{q}^T_{t-\delta t} \cdot \mathbf{g}}, \label{equNewRatio2}
\end{align}
let
\begin{align}
\underline{\lambda}_t=\min_i \frac{[\mathbf{q}^T_{t-\delta t}\mathbf{Q}^{\delta t} ]_i}{[\mathbf{q}^T_{t-\delta t} ]_i}, &&\overline{\lambda}_t=\max_i \frac{[\mathbf{q}^T_{t-\delta t} \mathbf{Q}^{\delta t} ]_i}{[\mathbf{q}^T_{t-\delta t} ]_i},
\end{align}
where $[\mathbf{v}]_i$ represents the $i$th entry in vector $\mathbf{v}$.

According to Collatz formula~\cite{collatz1942}\cite[Theorem 2]{wood2007finding}, 
\begin{align}
\lim_{t \to +\infty} \underline{\lambda}_t =\lim_{t \to +\infty}\overline{\lambda}_t =\rho(\mathbf{Q}^{\delta t} ).
\end{align} 
Hence for any $[ \mathbf{g}]_i >0$, it holds
\begin{align}
\lim_{t \to +\infty} \min_i \frac{[\mathbf{q}^T_{t-\delta t} \mathbf{Q}^{\delta t} ]_i [ \mathbf{g}]_i}{[\mathbf{q}^T_{t-\delta t} ]_i [ \mathbf{g}]_i}= \rho(\mathbf{Q}^{\delta t} ), &&
\lim_{t \to +\infty} \max_i \frac{[\mathbf{q}^T_{t-\delta t} \mathbf{Q}^{\delta t} ]_i [ \mathbf{g}]_i}{[\mathbf{q}^T_{t-\delta t} ]_i [ \mathbf{g}]_i}= \rho(\mathbf{Q}^{\delta t} ).
\end{align}
Using $\min \{ \frac{a_1}{b_1}, \frac{a_2}{b_2}\} \le \frac{a_1+a_2}{b_1+b_2} \le \max \{ \frac{a_1}{b_1}, \frac{a_2}{b_2}\}$, we get
\begin{align}
 \lim_{t \to +\infty}  \frac{\sum_i [\mathbf{q}^T_{t-\delta t} \mathbf{Q}^{\delta t} ]_i [  \mathbf{g} ]_i}{\sum_i[\mathbf{q}^T_{t-\delta t} ]_i [ \mathbf{g}]_i}= \rho(\mathbf{Q}^{\delta t} ).
\end{align}
Equivalently
\begin{align}
 \lim_{t \to +\infty} \frac{\mathbf{q}^T_{t-\delta t} \mathbf{Q}^{\delta t}    \mathbf{g}}{\mathbf{q}^T_{t-\delta t}  \cdot \mathbf{g}}= \rho(\mathbf{Q}^{\delta t} ).
\end{align}
Then
\begin{align}
 \lim_{t \to +\infty}  \left|\frac{f_{t+\delta t}-f_t}{f_t -f_{t-\delta t}}\right|^{1/\delta t} =\rho(\mathbf{Q}^{\delta t} )^{1/\delta t} =\rho(\mathbf{Q}).
\end{align}
Finally
it comes to the second conclusion.
\end{IEEEproof}

The theorem shows that the average convergence rate $R^\ddagger(t)$ plays the same role as   $R(t)$.  But the calculation of $R^\ddagger(t)$  is not as stable as that of $R(t)$  in practice. 

\section{Conclusions}
\label{secConclusions}
This paper proposes a new   convergence rate of EAs, called the average (geometric) convergence rate.  The rate represents a normalized geometric mean of the reduction ratio of the fitness difference  per generation.   The calculation of the average convergence rate is  simple and   easy to implement on most   EAs in practice. 
Since the  rate is normalized, it is convenient to compare   different EAs on optimization  problems.
 
For discrete optimization, lower   bounds on the average convergence rate of EAs have been established. It is proven that under random initialization,  the average convergence rate $R(t)$ for $t$ generations converges to a limit, called the asymptotic average convergence rate; and   under particular initialization, $R(t)$ equals to the asymptotic average convergence rate  for any $t \ge 1$.

The analysis of EAs for continuous optimization is different from that for discrete optimization. In continuous optimization, an EA is modeled by a Markov chain  on a general state space, rather than a finite Markov chain. So a different theoretical analysis is needed, rather than matrix analysis used in the current paper.  This topic is left for future research.  
% \bibliographystyle{IEEEtran}
% \bibliography{IEEEabrv,../../hejun-bib/hejun-2015,../../hejun-bib/hejunpl-2015,paper-A9}

\begin{thebibliography}{10}
\providecommand{\url}[1]{#1}
\csname url@samestyle\endcsname
\providecommand{\newblock}{\relax}
\providecommand{\bibinfo}[2]{#2}
\providecommand{\BIBentrySTDinterwordspacing}{\spaceskip=0pt\relax}
\providecommand{\BIBentryALTinterwordstretchfactor}{4}
\providecommand{\BIBentryALTinterwordspacing}{\spaceskip=\fontdimen2\font plus
\BIBentryALTinterwordstretchfactor\fontdimen3\font minus
  \fontdimen4\font\relax}
\providecommand{\BIBforeignlanguage}[2]{{%
\expandafter\ifx\csname l@#1\endcsname\relax
\typeout{** WARNING: IEEEtran.bst: No hyphenation pattern has been}%
\typeout{** loaded for the language `#1'. Using the pattern for}%
\typeout{** the default language instead.}%
\else
\language=\csname l@#1\endcsname
\fi
#2}}
\providecommand{\BIBdecl}{\relax}
\BIBdecl

\bibitem{ming2006convergence}
L.~Ming, Y.~Wang, and Y.-M. Cheung, ``On convergence rate of a class of genetic
  algorithms,'' in \emph{Proceedings of 2006 World Automation Congress}.\hskip
  1em plus 0.5em minus 0.4em\relax IEEE, 2006, pp. 1--6.

\bibitem{suzuki1995markov}
J.~Suzuki, ``A {Markov} chain analysis on simple genetic algorithms,''
  \emph{IEEE Transactions on Systems, Man and Cybernetics}, vol.~25, no.~4, pp.
  655--659, 1995.

\bibitem{schmitt2001importance}
F.~Schmitt and F.~Rothlauf, ``On the importance of the second largest
  eigenvalue on the convergence rate of genetic algorithms,'' in
  \emph{Proceedings of 2001 Genetic and Evolutionary Computation Conference},
  H.~Beyer, E.~Cantu-Paz, D.~Goldberg, Parmee, L.~Spector, and D.~Whitley,
  Eds.\hskip 1em plus 0.5em minus 0.4em\relax Morgan Kaufmann Publishers, 2001,
  pp. 559--564.

\bibitem{he1999convergence}
J.~He and L.~Kang, ``On the convergence rate of genetic algorithms,''
  \emph{Theoretical Computer Science}, vol. 229, no. 1-2, pp. 23--39, 1999.

\bibitem{rudolph1997local}
G.~Rudolph, ``Local convergence rates of simple evolutionary algorithms with
  {Cauchy} mutations,'' \emph{IEEE Transactions on Evolutionary Computation},
  vol.~1, no.~4, pp. 249--258, 1997.

\bibitem{rudolph1997convergence}
------, ``Convergence rates of evolutionary algorithms for a class of convex
  objective functions,'' \emph{Control and Cybernetics}, vol.~26, pp. 375--390,
  1997.

\bibitem{rudolph2013convergence}
------, ``Convergence rates of evolutionary algorithms for quadratic convex
  functions with rank-deficient hessian,'' in \emph{Adaptive and Natural
  Computing Algorithms}.\hskip 1em plus 0.5em minus 0.4em\relax Springer, 2013,
  pp. 151--160.

\bibitem{teytaud2006ultimate}
O.~Teytaud, S.~Gelly, and J.~Mary, ``On the ultimate convergence rates for
  isotropic algorithms and the best choices among various forms of isotropy,''
  in \emph{Parallel Problem Solving from Nature (PPSN IX)}.\hskip 1em plus
  0.5em minus 0.4em\relax Springer, 2006, pp. 32--41.

\bibitem{du2010convergence}
X.~Du and L.~Ding, ``About the convergence rates of a class of gene expression
  programming,'' \emph{Science China Information Sciences}, vol.~53, no.~4, pp.
  715--728, 2010.

\bibitem{beume2011convergence}
N.~Beume, M.~Laumanns, and G.~Rudolph, ``Convergence rates of {SMS-EMOA} on
  continuous bi-objective problem classes,'' in \emph{Proceedings of the 11th
  Workshop on Foundations of Genetic Algorithms}.\hskip 1em plus 0.5em minus
  0.4em\relax ACM, 2011, pp. 243--252.

\bibitem{jebalia2010log}
M.~Jebalia and A.~Auger, ``Log-linear convergence of the scale-invariant
  ($\mu/\mu_w$, $\lambda$)-{ES} and optimal $\mu$ for intermediate
  recombination for large population sizes,'' in \emph{Parallel Problem Solving
  from Nature (PPSN XI)}.\hskip 1em plus 0.5em minus 0.4em\relax Springer,
  2010, pp. 52--62.

\bibitem{teytaud2014convergence}
F.~Teytaud and O.~Teytaud, ``Convergence rates of evolutionary algorithms and
  parallel evolutionary algorithms,'' in \emph{Theory and Principled Methods
  for the Design of Metaheuristics}.\hskip 1em plus 0.5em minus 0.4em\relax
  Springer, 2014, pp. 25--39.

\bibitem{he2003towards}
J.~He and X.~Yao, ``Towards an analytic framework for analysing the computation
  time of evolutionary algorithms,'' \emph{Artificial Intelligence}, vol. 145,
  no. 1-2, pp. 59--97, 2003.

\bibitem{varga2009matrix}
R.~Varga, \emph{Matrix Iterative Analysis}.\hskip 1em plus 0.5em minus
  0.4em\relax Springer, 2009.

\bibitem{he2009mixed}
J.~He and L.~Kang, ``A mixed strategy of combining evolutionary algorithms with
  multigrid methods,'' \emph{International Journal of Computer Mathematics},
  vol.~86, no.~5, pp. 837--849, 2009.

\bibitem{yao1999evolutionary}
X.~Yao, Y.~Liu, and G.~Lin, ``Evolutionary programming made faster,''
  \emph{IEEE Transactions on Evolutionary Computation}, vol.~3, no.~2, pp.
  82--102, 1999.

\bibitem{salomon1998evolutionary}
R.~Salomon, ``Evolutionary algorithms and gradient search: similarities and
  differences,'' \emph{IEEE Transactions on Evolutionary Computation}, vol.~2,
  no.~2, pp. 45--55, 1998.

\bibitem{meyer2000matrix}
C.~Meyer, \emph{Matrix Analysis and Applied Linear Algebra}.\hskip 1em plus
  0.5em minus 0.4em\relax SIAM, 2000.

\bibitem{he2012pure}
J.~He, F.~He, and H.~Dong, ``Pure strategy or mixed strategy?'' in
  \emph{Evolutionary Computation in Combinatorial Optimization}, J.-K. Hao and
  M.~Middendorf, Eds.\hskip 1em plus 0.5em minus 0.4em\relax Springer, 2012,
  pp. 218--229.

\bibitem{jansen2014performance}
T.~Jansen and C.~Zarges, ``Performance analysis of randomised search heuristics
  operating with a fixed budget,'' \emph{Theoretical Computer Science}, vol.
  545, pp. 39--58, 2014.

\bibitem{grinstead1997introduction}
C.~Grinstead and J.~Snell, \emph{Introduction to Probability}.\hskip 1em plus
  0.5em minus 0.4em\relax American Mathematical Society, 1997.

\bibitem{collatz1942}
L.~Collatz, ``Einschlie{\ss}ungss\"atze f{\"u}r charakteristische zahlen von
  matrizen,'' \emph{Mathematische Zeitschrift}, vol.~48, no.~1, pp. 221--226,
  1942.

\bibitem{wood2007finding}
R.~J. Wood and M.~O'Neill, ``Finding the spectral radius of a large sparse
  non-negative matrix,'' \emph{ANZIAM Journal}, vol.~48, pp. 330--345, 2007.

\end{thebibliography}


\end{document}